\documentclass[conference]{IEEEtran}
\ifCLASSINFOpdf
  \usepackage[pdftex]{graphicx}
\else

\fi

\usepackage{amsmath}
\begin{document}
\title{Geometry-Aware Fruit Grasping Estimation for Robotic Harvesting in Orchards}

\author{\IEEEauthorblockN{Hanwen Kang$^\dagger$}
\IEEEauthorblockA{Department of\\Areospace and Mechanical\\
Monash University\\
Melboure, Australian\\}
\and

\IEEEauthorblockN{Xing Wang$^\dagger$}
\IEEEauthorblockA{Department of\\Areospace and Mechanical\\
Monash University\\
Melboure, Australian\\}
\and

\IEEEauthorblockN{Chao Chen}
\IEEEauthorblockA{Department of\\Areospace and Mechanical\\
Monash University\\
Melboure, Australian\\}}

\maketitle

\begin{abstract}
Field robotic harvesting is a promising technique in recent development of agricultural industry. It is vital for robots to recognise and localise fruits before the harvesting in natural orchards. However, the workspace of harvesting robots in orchards is complex: many fruits are occluded by branches and leaves. It is important to estimate a proper grasping pose for each fruit before performing the manipulation. In this study, a geometry-aware network, A3N, is proposed to perform end-to-end instance segmentation and grasping estimation using both color and geometry sensory data from a RGB-D camera. Besides, workspace geometry modelling is applied to assist the robotic manipulation. Moreover, we implement a global-to-local scanning strategy, which enables robots to accurately recognise and retrieve fruits in field environments with two consumer-level RGB-D cameras. We also evaluate the accuracy and robustness of proposed network comprehensively in experiments. The experimental results show that A3N achieves 0.873 on instance segmentation accuracy, with an average computation time of 35 ms. The average accuracy of grasping estimation is 0.61 cm and 4.8$^{\circ}$ in centre and orientation, respectively. Overall, the robotic system that utilizes the global-to-local scanning and A3N, achieves success rate of harvesting ranging from 70\% - 85\% in field harvesting experiments.
\end{abstract}

\IEEEpeerreviewmaketitle

\section{Introduction}
\begin{figure}[ht]
    \centering
    \includegraphics[width=.45\textwidth]{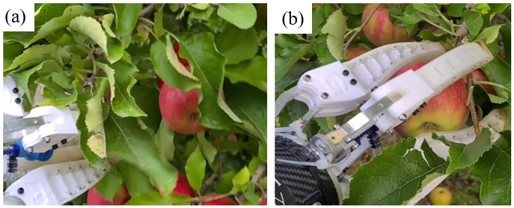}
    \caption{(a) Target apple surrounded by leaves and branches, (b) attempt to grasp the target apple with soft end-effector.}
    \label{fig:workspace}
\end{figure}

With the continuously increasing cost of the labor force, robotic fruit retrieving in orchards has become a promising technology in the near future \cite{li2011review,sarig1993robotics}. However, robotic fruit harvesting in common orchards' environments is more challenging than the traditional crop harvesting \cite{zhao2016review,font2014proposal}, because most of the orchards' environments are highly unstructured and complex. Therefore, most recent robotic fruit retrieving systems detach fruits from plants by applying an end-effector on the high Degree-of-Freedom (DoF) robotic arms \cite{zhou2021intelligent, wang2020soft}. In general, a fruit picking cycle includes four steps: perception, approaching, detachment, and collection. Perception by vision techniques is key to the success of robotic fruit harvesting, as robots need to see fruits before further processing \cite{gongal2015sensors}. After that, robots need to find a proper grasping orientation and a collision-free path to approach and detach fruits from trees. In the past several years, many methods have been developed for visual perception. Both traditional and deep-learning based methods are used to detect, segment, and localise fruits using the sensor data, such as RGB images\cite{chen2021deep,kang2020fast}, point clouds \cite{lin2020color}, etc. Most of the studies do not consider estimating approaching orientation of fruits. If the workspace is clear and fruits are not blocked by obstacles, it is not challenging to detach these fruits off trees. However, in many cases, the fruits are surrounded by tangled branches and leaves, as shown in Figure \ref{fig:workspace}(a) and (b). It is highly possible that the end-effector would fail to grasp fruits. A forced pulling back action of the end-effector may cause damage to both the robotic arm and trees. Meanwhile, in practice, the collision between robotic' bodies and environments may cause unexpected moving of target fruits, thus leading to a failure of harvesting cycle. Up to date, only a few studies \cite{guo2020pose,li2018pose,lin2019pose,eizentals2016pose,wang2021bio} have tried to solve this problem to improve the success rate of the harvesting. Besides, these works focus on fruit grasping estimation in structured environments, such as greenhouses and laboratories, and cannot be generalized to complex field environments \cite{kang2020grasping,kang2020visual}.

Motivated by the object pose estimation using multi-source data of RGB image and point cloud \cite{qi2018frustum}, a new geometry-aware deep-learning network model, Apple 3D Network (A3N), is proposed in this work. A3N is designed to perform end-to-end detection, instance segmentation, and grasping estimation of fruits using the raw RGB image and point clouds from an RGB-D camera. To be specific, A3N takes advantage of the deep-learning detector to search Region of Interest (ROI) from RGB images. Then, a PointNet model is utilised to perform the bounding boxes regression on point clouds of each fruit, which predicts a proper angle for robotic arm approaching using point clouds. Based on an estimated grasping pose of each fruit, the robot can plan a proper path to approach and detach fruit accurately and safely. To provide the robot with more information about surrounding obstacles, OctoMap \cite{hornung2013octomap} is used to construct the occupancy map of the workspace. Finally, the A3N is evaluated with the field data and our developed robotic retrieving system. To summarize, we make the following contributions in this research:
\begin{itemize}
    \item A novel end-to-end geometry-aware network A3N including a region proposal and a grasping estimation network is proposed to perform fruit segmentation and grasping pose estimation, respectively. 
    \item A framework including fruit detection, grasping pose estimation, and workspace modeling is proposed, which can be directly used for accurate and robust robotic harvesting in orchards.
    \item A global-to-local strategy is implemented, allowing accurate vision sensing in orchards when depth 
    sensors have limited accuracy. Extensive quantitative evaluations are also included to validate the proposed method.
\end{itemize}

The rest of the paper is organised as follows. Section \ref{section:related_work} reviews the related works. Sections \ref{section:System Design and Operation} introduces the design of the robotic system and the visual processing approach, respectively. Experimental methods and results are presented in Section \ref{section:Experiment and Discussion}. In Section \ref{section:Conclusion}, conclusion and future work are included.

\section{Related Works} \label{section:related_work}
Fruit detection is an essential step in robotic harvesting \cite{voulodimos2018deep-vs-review}. Traditional fruit detection uses handcrafted features, such as gradients and textures, to perform segmentation or detection \cite{blehm2005computer}. However, the performance of traditional method is limited in field environments \cite{kamilaris2018deep}. Comparatively, deep-learning method shows superior performance in accuracy and generalization \cite{kang2020fast}. Deep-learning based fruit detection includes two-stage \cite{gao2020multi} and one-stage methods \cite{chen2021deep}. Two-stage detection networks first search ROIs of objects, then perform the classification and location regression on ROIs \cite{ren2015faster-rcnn}. While one-stage detection networks directly perform the region proposal, classification, and regression in one step \cite{simony2018yolo}. Deep-learning based fruit detection has been widely studied in many scenes. Yu et al. (2019) applied mask-RCNN to detect strawberries in greenhouses, an average precision of 95.78 \% was reported \cite{yu2019strawberry}. Tian et al. (2019) applied YOLO-V3 to monitor apples' maturity and growing stage in the orchard environments \cite{tian2019apple}. Kang et al.(2019) presented a one-stage panoptic segmentation network to perform vision sensing in orchard conditions; an accuracy of 0.87 was reported \cite{kang2020fruit}.

Grasping estimation by the vision system can provide critical information for the manipulation \cite{lehnert2016grasping}. It has been widely studied in many robotic and computer vision tasks. Traditional grasping estimation utilize point cloud alignment methods, such as ICP \cite{droeschel2013icp-pose}, which match objects' shapes with templates. These methods suffer the performance degeneration when there are significant variances of the shape. Grasping estimation using deep-learning method has drawn tremendous attention recently. It uses the deep-learning algorithm to estimate a grasping pose with 2D and 3D data \cite{du2021grasping}. Grasping estimation using 2D images recasts the task as an image detection or classification problem \cite{zhu20142dgrasping}. While grasping estimation using 2D images can apply 3D data, such as points\cite{qi2017pointnet,fang2020graspnet}, voxel\cite{zhou2018voxelnet}, and meshes. However, only a few studies focus on grasping estimation in the robotic harvesting. Most of these existing works try to solve the grasping estimation of fruits in greenhouses \cite{eizentals2016pose,li2018pose,barth2019grasping_angle} or in factory lines \cite{guo2020pose}. Without proper vision information, the success rate of harvesting can be severely affected when robots work in fields \cite{hayashi2010evaluation}. Besides, some work \cite{lin2019pose,kang2020visual} has studied grasping estimation on fruits in fields while ignoring the collision objects around the fruits, which cannot ensure the safety and accuracy of the manipulation. With the help of visual perception, robotic harvesting is implemented by manipulators toward the human-like robotic harvesting \cite{oliveira2021advances}. Birrell et al. \cite{birrell2020field} applied a 6-DoF arm and an end-effector with force monitoring, achieving 82\% accuracy in clarifying vegetables. Ge et al. \cite{ge2019fruit} developed a dedicated manipulator to harvest strawberries in the greenhouse. A Mask-RCNN based vision system was used to localise fruits and model structured obstacles within the workspace; a 74.1\% success rate of harvesting on ripe strawberry was reported. Sepulveda et al. \cite{sepulveda2020robotic} applied a dual-arm system to harvest eggplant in the laboratory. One arm can remove occlusions, while another arm can pick fruits simultaneously. Even though these works used information from visual perception to harvest fruits and avoid collision between robots and environments during the operation, they did not consider vision perception and robotic operations in unstructured environments, which are the most common workspace for robotic operation in fruit orchards. Therefore, a vision perception algorithm that can perform grasping estimation and workspace modelling is urgent to the success of the robotic harvesting system in fields.

\section{Material and Methodology} \label{section:System Design and Operation}
\subsection{Data Collection}
The data were collected in apple orchards located in Melbourne, Australia, using a Realsense D435 camera. The time of data collection was between 10:00 am to 4:00 pm. The distance between cameras and trees was from 0.4 to 1.2 meters. In total, 768 RGB-D images and another 1132 color images were taken in orchards. The ground truth of the detection and segmentation were labelled using the software LabelMe. The ground truth of grasping poses was labelled using our dedicated developed software.

\subsection{Geometry-aware Grasping Estimation}
\begin{figure}[ht]
    \centering
    \includegraphics[width=.4\textwidth]{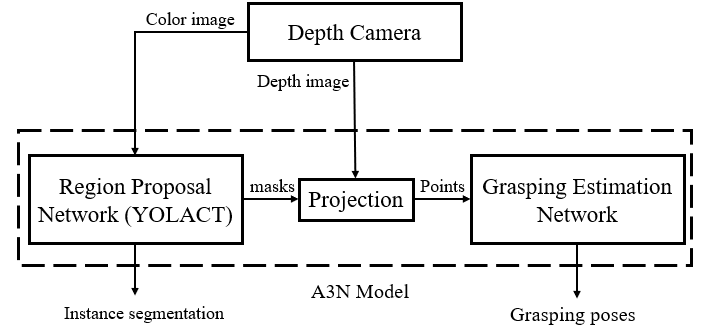}
    \caption{Working framework of the proposed A3N: Region Proposal Network receives and processes the RGB image, Grasping Estimation Network receives and processes the depth image filtered by generated masks.}
    \label{fig1:a3n-pipeline}
\end{figure}
A3N includes two modules: a region proposal network to process 2D images and a grasping estimation network to process point cloud, as shown in Figure \ref{fig1:a3n-pipeline}. Region proposal network applies a 2D detector to perform detection and instance segmentation of objects. Then a modified PointNet is used to perform grasping estimation using point cloud within the instance segmented and surround area of each object. The grasping poses of each fruit are returned as the output for manipulation.

\subsubsection{Region Proposal Network}
\begin{figure*}[ht]
    \centering
    \includegraphics[width=.8\textwidth]{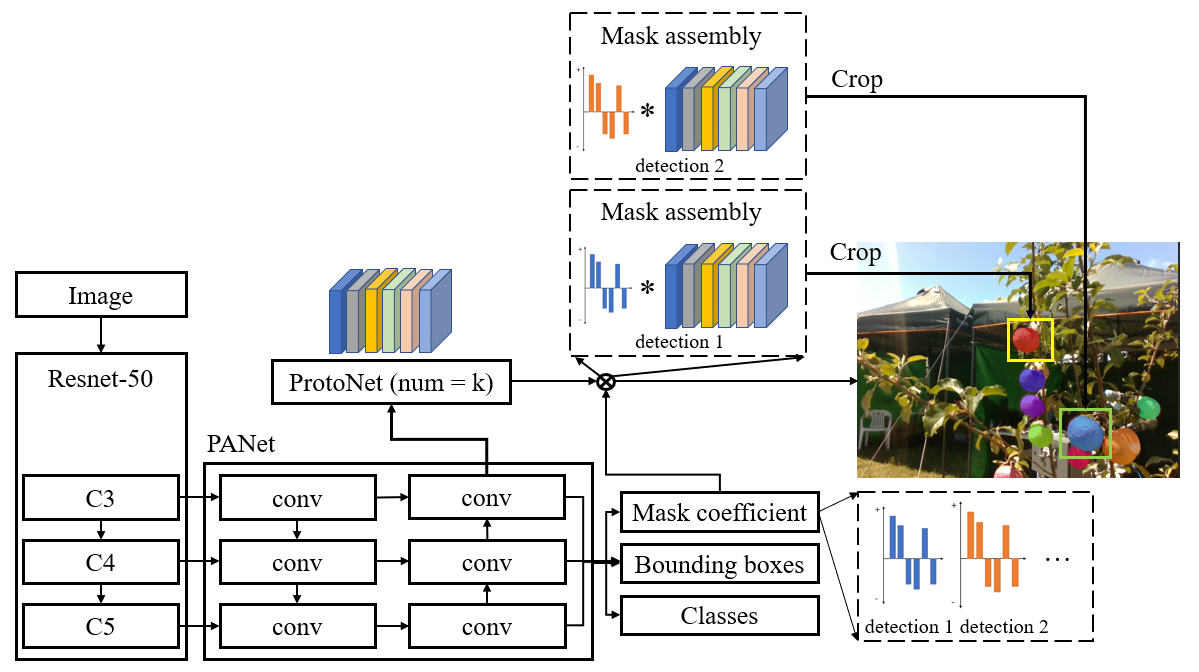}
    \caption{Architecture of the 2D region proposal network, which can extract ROI from RGB images. We base this architecture off of ResNet-50 + FPN (PANet) }
    \label{fig2:a3n-region-proposal}
\end{figure*}

RGB images have many dense features on objects' appearances compared to sparse 3D data. A3N takes advantage of 2D deep-learning methods, using a YOLACT model \cite{bolya2019yolact}, to perform instance segmentation in a one-stage detector without explicit feature localisation step. Figure \ref{fig2:a3n-region-proposal} shows the architecture of the model. YOLACT performs instance segmentation in two parallel tasks; one is to produce image-sized semantic masks in FCN \cite{long2015fully} way, another is to predict parameters to segment those semantic masks into the instance level explicitly. YOLACT network includes three branches: a backbone network, a detection branch, and an instance segmentation branch. We use ResNet-50 \cite{he2016deep} as the default backbone network, and the base image size is 416 $\times$ 416. The feature maps of C3, C4, and C5 levels in backbone network are output for further processing. The detection branch has two components: Feature Pyramid Network (FPN), and detection head. FPN uses PANet \cite{wang2019panet} to fuse multi-scale feature maps from the backbone. PANet receives feature maps from C3, C4, and C5 levels of the backbone and each level applies a detection head to predict detection. Each detection head has $1+4+c+k$ channels, which are parameter numbers of confidence score, box localisation, classification, and mask assembly, respectively.

\textbf{Mask Assembly}: The instance segmentation branch generates prototype masks using feature maps in size of the C3 level. The input feature maps can from the backbone network or PANet. The prototype masks are identical to semantic feature maps in FCN, whose last layer has $k$ channels. The instance masks are assembled using matrix multiplication of prototype masks (size is $h \times w \times k$) and mask coefficients (size is $n \times k$). Then the results are processed by sigmoid activation to produce final masks. The final masks are cropped into instance masks using the information of objects' bounding boxes from detection head.

\subsubsection{Grasping Estimation Network}
\begin{figure}[ht]
    \centering
    \includegraphics[width=.48\textwidth]{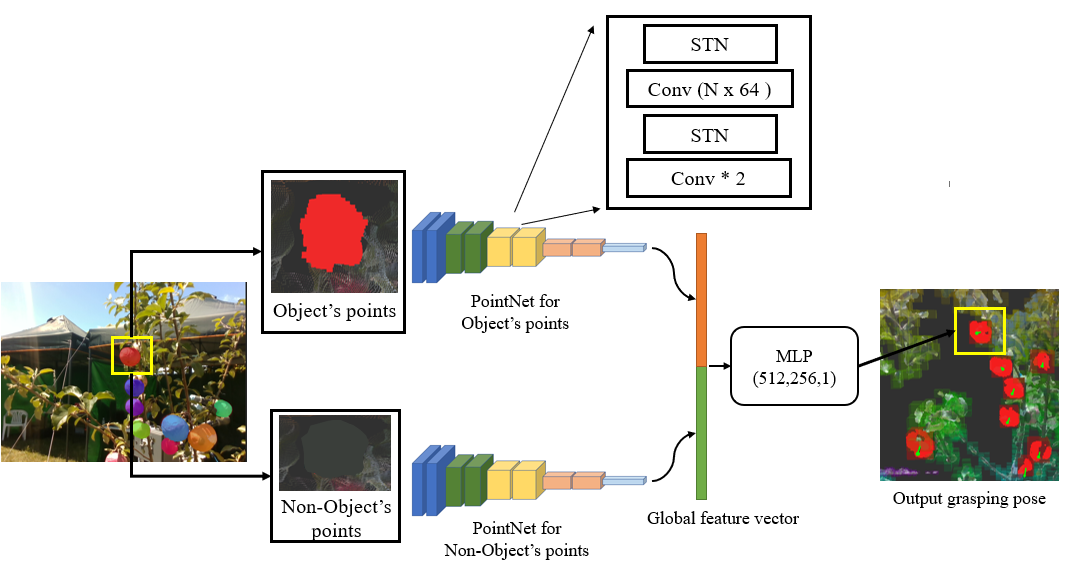}
    \caption{Working framework of the grasping estimation network: both objects' and non-objects' points are utilized in the separate subnet}
    \label{fig3:a3n-pipeline}
\end{figure}

From the previous step, a set of bounding boxes and instance masks of objects in image space are obtained. A grasping estimation network is then applied to estimate the grasping pose of each fruit using the PointNet-based network.

\textbf{PointNet architecture}:  Point set has two essential characters: order and transform invariance, which means the properties of a point cloud will not change with alteraction of point order or the object's pose. For the first property, PointNet uses a symmetric function to extract a feature vector invariant to point order. For the second property, PointNet uses Spatial Transform Net (STN) \cite{ja2015STN} to predict an affine transformation matrix on input raw points directly. Moreover, PointNet uses shared Multi-layer Perception (MLP) to aggregate geometry features in local and global scales. PointNet has enormous advantages in computation complexity as it enables 2D network architecture in 3D data learning without scarifies of data resolution. 

\textbf{Geometry-aware Grasping Estimation}: A modified PointNet is used to estimate the grasping pose of each fruit, as shown in Figure \ref{fig3:a3n-pipeline}. The modified model has two subnets to receive points of both objects and non-objects. Each subnet has five blocks: one STN followed by a convolution layer (MLP) to convert points set ($n \times 3$) into a feature vector with the size of $n \times 1 \times 64$, where $n$ is the number of points in the set. Then another STN is used to estimate a proper transform in feature space followed by another two convolution layers, which outputs a feature vector with the size of $n \times 1 \times m$ (m is set as 256). Lastly, a symmetric function, maximum pooling, is used on the first dimension of the feature vector, producing a feature vector in size of $1 \times 1 \times m$. The output two feature vector is resized into $m \times 1$ and concatenated into a $2m \times 1$ feature vector, named the global feature vector. After that, three fully connected layers is used to generate the grasping pose bounding box of each fruit.

\textbf{Grasping Pose Representation}: We use Euler-angle to represent the orientation of the grasping poses. To ensure the safety of the robot system, the value of predicted angle into the region of $[\frac{\pi}{4},-\frac{\pi}{4}]$. To ensure the convergence of network training, we align a local coordinate at the centre of each points set. Network estimates offsets on the X-,Y-,Z- axis to obtain the real location of objects' centre.

\textbf{Point Cloud processing}: 3D points set of instance masks can be computed using a camera projection matrix. The intrinsic and extrinsic matrices of the RGB camera are calibrated before implementation. The computed points are divided into two sets: object's point and non-object's points, based on the instance masks. For the non-object points, the points within the range of 0.3 m from the centre of objects are retained. Then, a voxel-sampling algorithm is applied to re-sample the points into a given resolution.

\subsubsection{Network Training}
We train YOLACT and grasping estimation network separately. Three losses are utilized to train YOLACT: classification loss $L_{cls}$, boxes regression loss $L_{box}$, and mask loss $L_{mask}$. Classification and box regression losses use binary cross-entropy as only two classes are involved, while mask loss uses L2 losses. The weight of the $L_{cls}$, $L_{box}$, and $L_{mask}$ are 1.0, 1.0, and 2.5, respectively. The training of the grasping estimation network includes box regression loss $L_{box}$ and grasping pose regression loss $L_{ori}$. Both $L_{box}$ and $L_{ori}$ use smooth-L1 (Huber) loss. The weight of the $L_{box}$ and $L_{ori}$ are 1.0 and 2.0, respectively. 

\subsection{Robotic Manipulations}
\begin{figure*}[ht]
    \centering
    \includegraphics[width=.8\textwidth]{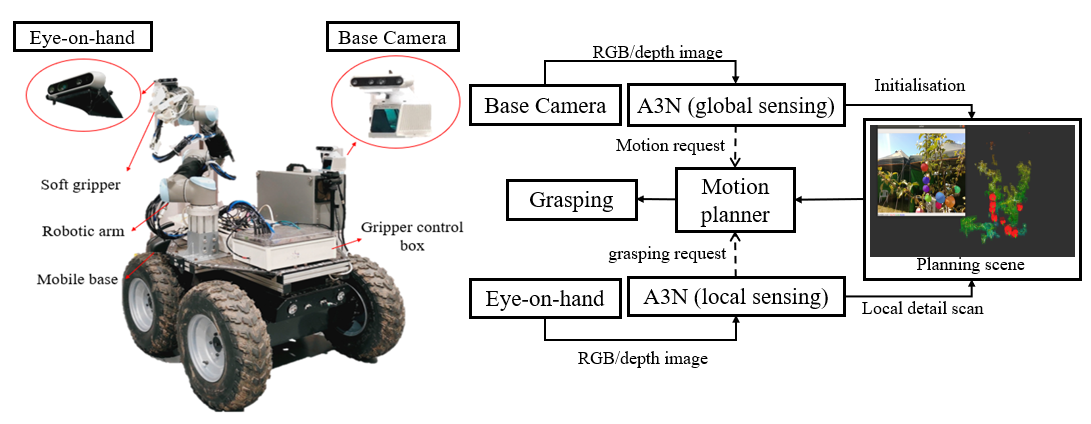}
    \caption{The robotic harvesting system integration with both base camera and eye-on-hand. Working framework of the global-to-local sensing strategy: the A3N performs global sensing first to obtain the position of targets and construct the scene of the workspace, which guides eye-on-hand local sensing to refine the detection and execute the grasping. }
    \label{fig:a3n-global-local}
\end{figure*}

The harvesting robot mainly includes a mobile base, an industry robotic arm (UR5), a vision system, and a soft end-effector, as shown in Figure \ref{fig:a3n-global-local}. The vision system includes two Intel RealSense RGB-D cameras: one at the base while the other is on the end-effector. Both cameras are connected to the central computer, an NVIDIA Jetson-TX2. The control framework is implemented on Robotic Operation System (ROS) melodic.

\textbf{Global-to-local strategy}: The harvesting system equipped with two RGB-D cameras on the base and the end-effector respectively, as shown in Figure \ref{fig:a3n-global-local}. Recent studies \cite{fu2020application,neupane2021evaluation} show that the accuracy of consumer RGB-D cameras, such as Realsense D435, will significantly degenerate when distance exceeds 1 meter. However, the field of view for the camera will be largely limited if the distance is too close. Therefore, a global-to-local sensing strategy is implemented here. The base camera is first used to scan the global fruits and initialise a raw workspace model for manipulation planning. Manipulator is then moved to each target fruit in consequence. Eye-on-hand camera is utilized to update the accurate local model. In such manner, workspace model and fruit poses are gradually refined, ensuring the accuracy and efficiency of harvesting.

\textbf{Workspace modelling}: Workspace modelling is an essential step for robotic operations in field environments as the collisions existed in environments can heavily affect the operation of the manipulation. Octomap \cite{hornung2013octomap}, which uses octree to subdivide the occupancy into voxel in given size hierarchically, is applied to turn collision points into occupancy grid (resolution set as 5cm). During the operation, joint state messages from the robot arm are used to register each frame of point cloud into the map. With a global-to-local sensing strategy, an Octomap with fine details is presented at last for manipulation planning.

\textbf{Manipulation}: MoveIt! framework \cite{chitta2016moveit} is used to plan and execute manipulation. In our case, Fruit poses are first mapped into configuring space using the inverse kinematic algorithm, Trac-IK \cite{beeson2015trac}. The planning pipeline of MoveIt!, which uses the RRT algorithm in OMPL \cite{2012ompl}, receives the latest workspace model and current robot state to plan a collision-free path to the pose goal. Lastly, time-optimal trajectory generation is applied to smooth the trajectory. The trajectory is implemented using the Universal ROS driver. The system will repeat the aforementioned steps until all fruits are retrieved in the current workspace.

\section{Experiment and Discussion} \label{section:Experiment and Discussion}
\subsection{Evaluation Methods}
We use the F1-score and intersection of Union (IoU) to evaluate the performance of the A3N on detection and instance segmentation, respectively \cite{kang2020fruit}. IoU measures the ratio of the intersection area between the prediction and ground truth. The IoU of detection and segmentation are donated as $IoU_{det}$ and $IoU_{seg}$, respectively. The detected objects with a confidence score and $IoU_{det}$ larger than 0.5 are considered as the true positive. The performance of the grasping estimation network is evaluated using Root Mean Squared Error (RMSE) on both the centre position (cm) and angles ($^\circ$).

\subsection{Evaluation on Instance Segmentation}
This section reports the instance segmentation accuracy and detection performance of region proposal network of A3N . We also evaluate different network configurations and compare them with other detectors.

\textbf{Implementation details}: All models are trained using data collected from the orchards. The network's backbone adopts pre-trained weights from ImageNet. In the training, we first froze backbone weights, trained the rest of the model for 80 epochs, then trained the whole network for another 40 epochs. Adam-optimiser is applied with a learning rate of $10^{-3}$. The decay rate of optimiser and batch-norm layers are set as 0.95 and 0.9, respectively. We use batch size of 24 on one GPU (11GB) in first 80 epochs and 12 on the rest of epochs. Each model is trained for three times, the weights with the highest validation accuracy is utilized in evaluation.

\textbf{Comparison on Configurations}: We evaluate the performance of the alternative network configuration of the region proposal network. Firstly, we train networks with ResNet-50 (R50), ResNet-101 (R101), and MobileNet-v2 (MN) as the backbone. Then, we compare the instance segmentation accuracy of network using features from C3 or PANet, respectively. Also, we evaluate the performance of the network with different input image resolutions. Lastly, we compare the accuracy of instance segmentation by using different mask coefficients in the detection head. The experimental results are shown in Tables \ref{table: comparison of config} and \ref{table: comparison of mask coefficient}.
\begin{table}[ht]
    \centering
    \caption{Comparison of the performance on different configurations of the Region Proposal Detector of A3N\\}
    \begin{tabular}{c c c c c c}
    \hline
    Model & Backbone & Time & F1 & $IoU_{mask}$ \\
    \hline
    A3N-416 & R50-C3 & 35 & 0.890 & 0.842\\
    A3N-416 & MN-PANet & \textbf{24} & 0.873 & 0.851\\
    A3N-416 & R50-PANet & 35 & 0.890 & 0.873\\
    A3N-480 & R50-PANet & 53 & 0.903 & 0.884\\
    A3N-640 & R50-PANet & 76 & 0.897 & 0.886\\
    A3N-416 & R101-PANet & 47 & 0.907 & 0.882\\
    A3N-480 & R101-PANet & 75 & \textbf{0.923} & 0.891\\
    A3N-640 & R101-PANet & 97 & \textbf{0.923} & \textbf{0.893}\\
    \hline
    \end{tabular}
    \label{table: comparison of config}
\end{table}

\begin{table}[ht]
    \centering
    \caption{Performance of instance segmentation under various $k$ values}
    \begin{tabular}{c c c}
    \hline
    k & $IoU_{mask}$ & Time \\
    \hline
    8 & 0.863 & \textbf{34ms}  \\
    16 & 0.868 & 34ms  \\
    32 & \textbf{0.873} & 35ms  \\
    64 & 0.872 & 37ms  \\
    128 & 0.870 & 40ms  \\
    \hline
    \end{tabular}
    \label{table: comparison of mask coefficient}
\end{table}

\textbf{Comparison with other models}:  We compare the region proposal network of A3N with  YOLO-V4 \cite{yolov4} and mask-RCNN \cite{he2017mask}. Compared to the YOLO-V4, our region proposal network, applies depth-wise convolution layers instead of standard convolution layers in PANet. We also optimize some other details of the network model accordingly - for example, the layer configuration in PANet and detection head. Besides, we compare the model with Mask-RCNN, which has SOTA accuracy on instance segmentation. Both YOLO-V4 and Mask-RCNN are trained on collected data by using COCO pre-trained weights. While the training parameters of each model are slightly adjusted based on results. The results are shown in Table \ref{table: comparison of model}.
\begin{table*}[ht]
    \centering
    \caption{Comparison of performance among A3N, YOLO-V4 and Mask-RCNN with different architectures.\\}
    \begin{tabular}{c c c c c c}
    \hline
    Model & Backbone & Time & Time(TX2) & F1 & $IoU_{mask}$ \\
    \hline
    YOLO-V4-416  & CSPD53-PANet & 78 & 592 & 0.864 & N/A  \\
    YOLO-V4-480  & CSPD53-PANet & 106 & 827 & 0.886 & N/A  \\
    Mask-RCNN-640 & R50-FPN & 122 & 920 & 0.857 & 0.887 \\
    Mask-RCNN-640 & R101-FPN & 157 & 1285 & 0.877 & \textbf{0.895} \\
    A3N-416 & MN-PANet & \textbf{24} & \textbf{174} & 0.873 & 0.851\\
    A3N-416$^{*}$ & R50-PANet & 35 & 282 & 0.890 & 0.873\\
    A3N-480 & R101-PANet & 75 & 598 & \textbf{0.923} & 0.891\\
    A3N-640 & R101-PANet & 97 & 782 &\textbf{0.923} & 0.893\\
    \hline
    \end{tabular}
    \label{table: comparison of model}
\end{table*}

The performance evaluations of alternative network configuration are shown in Tables \ref{table: comparison of config} and \ref{table: comparison of mask coefficient}. Firstly, results suggest that prototype masks generation using features from PANet can significantly improve the accuracy of instance segmentation. This is because PANet can fuse robust semantic features of deeper levels into the lower level, and achieve better performance. We also trained the network using different image sizes and backbones. Results suggest that network performance would increase with raising of image size or using a backbone with higher accuracy, while the inference speed is inevitably reduced, as expected. Lastly, we compare the network accuracy on the segmentation with different number of mask coefficients. Results suggest that the region proposal network achieves the best accuracy on instance segmentation when $k$ is 32.

\begin{figure}[ht]
    \centering
    \includegraphics[width=.45\textwidth]{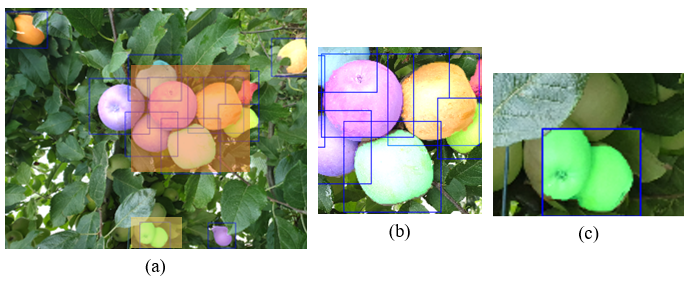}
    \caption{(a) Inaccuracy of instance segmentation results: (b) mask leakage and (c) inaccurate object localisation. }
    \label{fig: instance result}
\end{figure}

As shown in Table \ref{table: comparison of model}, region proposal network of A3N offers competitive accuracy on instance segmentation compared to the Mask-RCNN. The accuracy of A3N-640-R101-PANet achieves 0.893 on $IoU_{mask}$, which achieves equal performance compared to the Mask-RCNN. However, our model is mush faster in computational speed compared to the Mask-RCNN. From experiments, we observed two common reasons that lead to instance segmentation errors. The first reason is due to the mask leakage, which always occurs when fruits are close to each other, as shown in Figures \ref{fig: instance result}(a)(b). In this case, the network may fail to accurately segment boundary of each fruit. The second reason is the error of bounding box localisation, as shown in Figure \ref{fig: instance result} (c). Two apples are included in one bounding box, leading to the failure of the segmentation of each fruit. These two defects are more likely to occur when the sensing distance is larger. Therefore, an additive perception of fruits at a close distance can largely improve segmentation accuracy in our case.

\textbf{Computational Speed}: The inference time of networks is tested on an NVIDIA Jetson-TX2 and an NVIDIA RTX-2060 super. Table \ref{table: comparison of model} shows that A3N is faster than Mask-RCNN and YOLO-V4. Combining with network performance, our model similar or even better performance compared to the SOTA network but with faster speed in application of fruit detection and segmentation. Jetson TX2 is an embedded computer that widely applied in robotic applications. Our A3N-R50-PANet is $4.9\times$ faster compared to the Mask-RCNN on Jetson-TX2 with competitive accuracy. 

\subsection{Evaluation on Grasping Estimation}
\begin{figure}[ht]
    \centering
    \includegraphics[width=.40\textwidth]{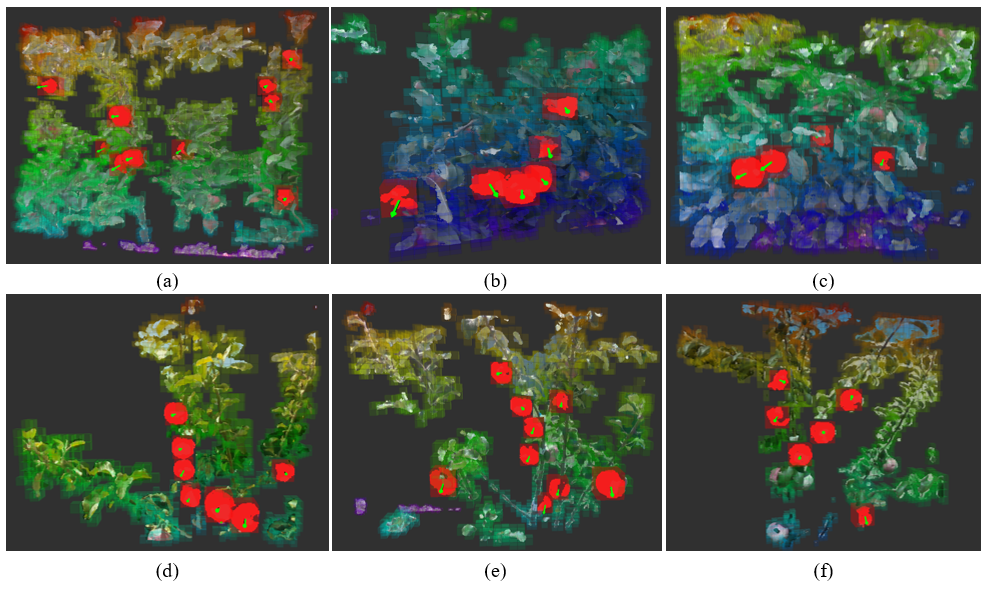}
    \caption{Planning scenes which include obstacles and grasping poses of the detected targets generated by the A3N and OctoMap. Red dots represented the targets and green arrow represents the suggested grasping orientation.}
    \label{fig6:a3n-planning_scene}
\end{figure}

\begin{figure}[ht]
    \centering
    \includegraphics[width=.45\textwidth]{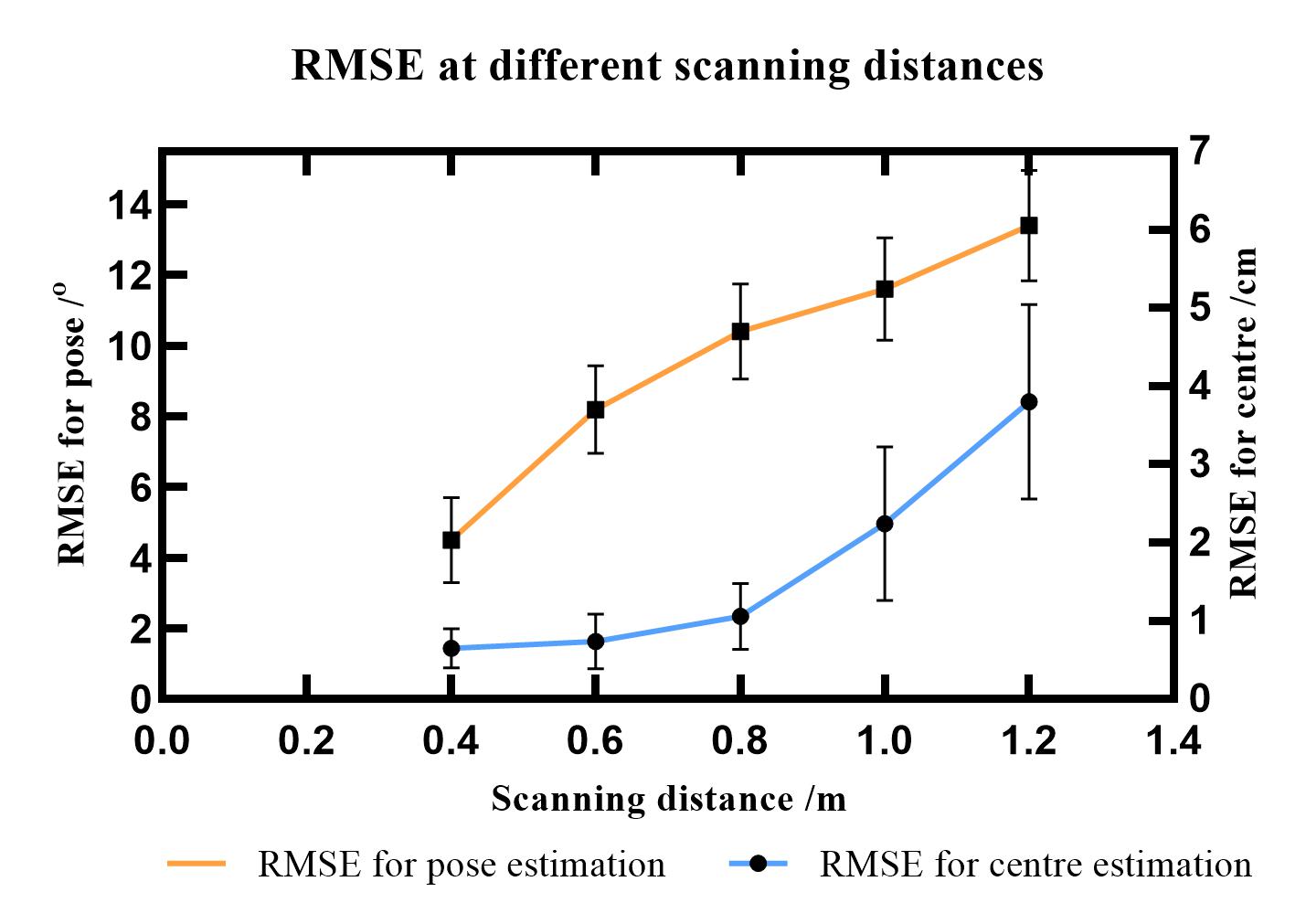}
    \caption{RMSE for both centre (cm) and pose estimation ($^\circ$) under different scanning distances. Note all the distances have an error range of $\pm$ 5cm.}
    \label{fig:error_vision}
\end{figure}

\begin{figure}[ht]
    \centering
    \includegraphics[width=.35\textwidth]{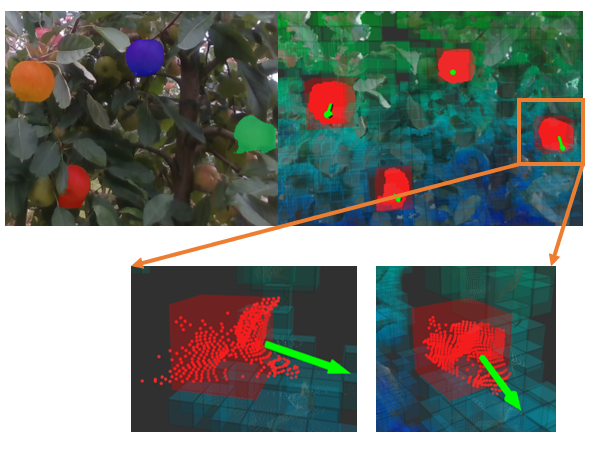}
    \caption{Illustration of data corruption in sensory data in field environments.}
    \label{fig:defect points}
\end{figure}

\textbf{Accuracy Test}: We evaluate the accuracy of A3N in fruits grasping estimation first. The instances of grasping estimation and reconstructed workspace are shown in Figure \ref{fig6:a3n-planning_scene}. The accuracy of grasping estimation is measured by RMSE between predicted pose with ground truth. The experimental results are shown in Figure \ref{fig:error_vision}. Based on the collected test dataset, we further analyze the grasping estimation performance according to the different distances that these images were taken. It can be seen that the centre and angular error of estimation are significantly reduced with a closer perception distance. Results also indicate that centre estimation achieves acceptable prediction accuracy within 0.8 m distance, while the grasping pose estimation can achieves accurate results when distance is about 0.4 m. With the increase of distance, the quality and number of points decrease dramatically. Meanwhile, sensory data always include defects (as shown in Figure \ref{fig:defect points}), which also affect the accuracy of estimation. From experiments, A3N can estimate fruit poses in fine accuracy in most cases. 
\begin{table*}[ht]
    \centering
    \caption{Comparison of inference time, RMSE of centre (cm) and angular estimation ($^\circ$) affected by the quality of instance segmentation generated by difference networks.\\}
    \begin{tabular}{c c c c c}
    \hline
    Model & Time & \multicolumn{2}{c}{ RMSE$_{centre}$} & RMSE$_{angular}$ \\
    \hline
     & TX2 & 0.4m  & 0.8m & 0.4m\\
     \hline
    A3N-416(MN) & 207 & 1.35 $\pm$ 0.41  & 1.8 $\pm$ 0.73 & 6.6 $\pm$ 3.1 \\
    A3N-416(C3) & 322 & 1.15 $\pm$ 0.37  & 1.7 $\pm$ 0.62 & 5.8 $\pm$ 2.5 \\
    A3N-416(PANet)$^{*}$ & 322 & 0.61 $\pm$ 0.25  & 1.06 $\pm$ 0.4 & 4.8 $\pm$ 2.2\\
    A3N-640(R101) & 816 & 0.53 $\pm$ 0.22  & 0.97 $\pm$ 0.35  & 4.4 $\pm$ 1.8 \\
    Mask-RCNN(R101)  & 1357  & 0.51 $\pm$ 0.22  & 0.99 $\pm$ 0.38  & 4.3 $\pm$ 1.6 \\
    \hline
    \end{tabular}
    \label{table: grasp on mask}
\end{table*}

\textbf{Influence of segmentation}: We report the influence of instance segmentation quality on grasping estimation in Table \ref{table: grasp on mask}. From the results, it can be seen that higher instance segmentation accuracy can improve grasping estimation results. The grasping estimation by using instance segmentation of Mask-RCNN achieves the highest accuracy with smaller variances. However, grasping estimation using A3N-416-PANet (the second row of Table \ref{table: grasp on mask}) also achieves competitive results with 4.2$\times$ faster speed. Therefore, we use the A3N-416-PANet as the region proposal network to compensate for the accuracy and the speed.

\begin{table*}[ht]
    \centering
    \caption{Comparison of performance among different input corruptions\\}
    \begin{tabular}{c | c | c c c | c c c}
    \hline
    Distance &  &\multicolumn{3}{c}{Missing} & \multicolumn{3}{|c}{Outlier}\\
    \hline
    Centre & raw (cm) &$10\%$ & $20\%$ & $40\%$ & $10\%$ & $20\%$ & $40\%$\\
    \hline
    0.4 m & 0.61 & 0.67 & 0.94 & 1.4 & 0.65 & 0.69 & 0.82 \\
    0.8 m & 1.06 & 1.22 & 1.48 & 1.97 & 1.16 & 1.27 & 1.43\\
    \hline
    Angular & raw ($^{\circ}$) & & &  &  &  &\\
    \hline
    0.4 m & 4.8 & 5.2 & 6.6 & 10.4 & 5.1 & 6.2 & 8.6\\
    0.8 m & 10.5 & 11.5 & 14.2 & 20.6 & 11.3 & 13.7 & 18.5\\
    \hline
    \end{tabular}
    \label{table: grasp on input}
\end{table*}

\textbf{Robustness Test}: To evaluate the robustness of proposed algorithm, we test A3N accuracy under various input corruptions, as shown in Table \ref{table: grasp on input}. From the results, grasping estimation has higher robustness on fruit centre prediction than grasping pose estimation. At the distance of 0.8 meter, A3N could still achieve high accuracy on fruit centre estimation when there are 40$\%$ point missing or 40$\%$ outliers points. This can make sure that robotic arm moves accurately to a close position to perform another precise perception. Grasping pose estimation at 0.4 meter shows the high robustness to the missing points or outliers, which enables robot to detach fruits in a proper orientation.

\subsection{Evaluation of A3N in Harvesting}

\begin{figure}[ht]
    \centering
    \includegraphics[width=.45\textwidth]{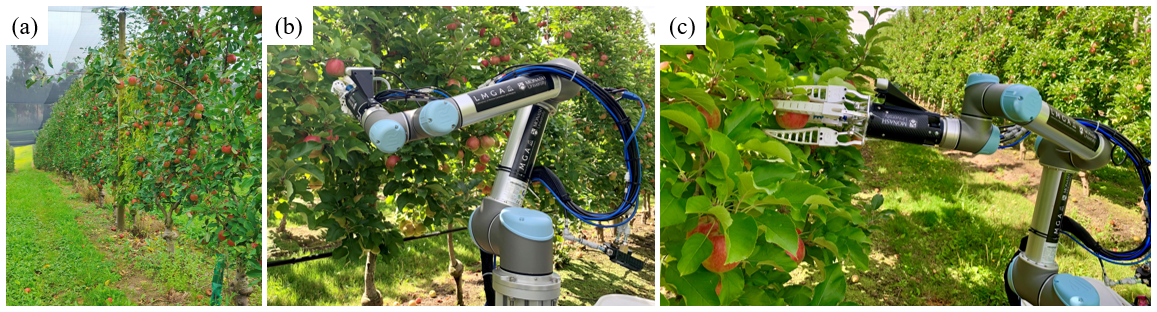}
    \caption{(a) Workspace of Fankhauser apple farm in Melbourne, (b) local sensing with A3N to refine the detection and workspace reconstruction, (c) grasping of target apple with a compliant end-effector.}
    \label{fig:fankhauser}
\end{figure}
This experiment evaluates the implementation performance of A3N in operation. The developed retrieving robotic system is tested in apple orchard in Melbourne, as shown in Figure \ref{fig:fankhauser}. There are other errors existing due to the system integration, such as the calibration error between eye (camera) and arm, manipulation error, or sensing error due to depth camera, etc. Therefore, it is difficult to independently evaluate the accuracy of the vision system. In the test, we move the manipulator to the estimated pose of fruits based on the perception results. Then, we manually measure the error between the gripper to the real pose of the fruits. We separately evaluate the robotic system by only using global, local, and global-to-local scanning. The results are shown in Table \ref{tab:t3}.
\begin{table}[ht]
    \centering
    \caption{Comparison among average experimental results of global, local, and global-to-local scanning strategy, while Num$_{Obs}$ represents the number of objects detected, Suc rate represents the successful rate of apple harvesting.}
    \begin{tabular}{ccccc}
    \hline
     Method & Num$_{Obs}$ & RMSE$_{ori}$($^{\circ}$) & RMSE$_{centre}$(cm) & Suc rate $\%$\\
     \hline
     Global & 16 & 13.7 & 3.9 & 56.3\\
     Local & 5 & 6.1 & 1.4 & 80\\
     G-to-L & 16 & 6.5 & 1.5 & 81.3\\
    \hline
    \end{tabular}
    \label{tab:t3}
\end{table}

As shown in results, both centre and angular error of the whole system are in the tolerance range of the gripper operation (maximum 3cm). By global scanning, the number of fruit detected by perception is maximized within the view. With following local perception, the accuracy of grasping estimation of each fruits are significantly improved. The system with local scanning alone requires multiple scans by manipulating to different locations, which lowers the efficiency of the system. Our robotic system achieves an average harvesting success rate from 70 - 85\% in operation, depending on the complexity of workspace. 

\section{Conclusion and future work} \label{section:Conclusion}
This work presents a geometry-aware detection network for apple harvesting applications, including the region proposal and grasping estimation networks. The former performs the fruit detection and instance segmentation tasks, while the latter predicts the 3D boundary box, centre of the fruit, and the appropriate approaching angle for grasping. Fruit detection and instance segmentation follow the work of YOLACT that combines these two tasks within one step, while pose estimation is developed and improved based on the PointNet. With the proposed A3N network, an F1-score of 0.890 is achieved for apple detection, an $IoU_{seg}$ of 0.873 is recorded for instance apple segmentation. The global and local scanning strategy achieves the RMSE of 0.61 cm for centre estimation and 4.8 $^\circ$ for angle estimation. Finally, a global-to-local scanning strategy is proposed and experimentally validated, which provides valuable guidance for the robot. An overall harvesting rate of 70\% - 85 \% is achieved on various natural orchard scenes.   

There are still many challenges remaining for current field robotic harvesting system. For example, the accuracy of RGB-D camera is significantly effected by sunlight. Potential solution such as Lidar, can be used to improve accuracy of perception from source. Besides, the path planning of the arm can also be improved to generate more humanoid behavior in grasping. Such improvements are expected to significantly improve the efficiency and success rate of robotic harvesting in field.

\section*{Acknowledgement}
We gratefully acknowledge the financial support from Australian Research Council (ARC ITRH IH150100006).
We would also like to thank Dr. Shao Liu and Dr Lilian Khaw at Monash University for language check.

\bibliography{root}

\end{document}